%% file: main.tex
\documentclass[sigconf]{acmart}

\usepackage{graphicx}
\usepackage{multirow}
\usepackage{tabularx}
\usepackage{enumitem}
\usepackage{dblfloatfix}
\usepackage{float}
\usepackage{xcolor}
\usepackage{soul}
\usepackage[normalem]{ulem}

\newcommand{\University}{\revise{an anonymous university}{the University of Wisconsin-Madison}}
\newcommand{\system}{\texttt{SHRIMP}}

\newcommand{\cOne}{\texttt{No-Plan}}
\newcommand{\cTwo}{\texttt{Plan-Only}}
\newcommand{\cThree}{\texttt{Plan+Edit}}

\newcommand{\tOne}{\texttt{Set the Table}}
\newcommand{\tTwo}{\texttt{Prepare a Snack}}
\newcommand{\tThree}{\texttt{Clean the Table}}

\newcommand{\sOne}{\texttt{Stepwise}}
\newcommand{\sTwo}{\texttt{Cumulative}}
\newcommand{\sThree}{\texttt{Oneshot}}

\newif\ifCOMMENTS
\COMMENTSfalse
\ifCOMMENTS
\newcommand{\revise}[2]{\textcolor{red}{\sout{#1}}\hl{#2}} % this strikes out old text and highlights new text
\else
\newcommand{\revise}[2]{#2}
\fi

\AtBeginDocument{%
  }

\copyrightyear{2026}
\acmYear{2026}
\setcopyright{cc}
\setcctype{by}
\acmConference[UIST '26]{The 39th Annual ACM Symposium on User Interface Software and Technology}{November 02--05, 2026}{Detroit, MI, USA}
\acmBooktitle{The 39th Annual ACM Symposium on User Interface Software and Technology (UIST '26), November 02--05, 2026, Detroit, MI, USA}
\acmDOI{10.1145/3830398.3830644}
\acmISBN{979-8-4007-2856-3/2026/11}

\begin{document}

\title{\system : Iterative Refinement of Robot Task Plans}

\author{Mya Schroder}
\orcid{0009-0005-9718-2684}
\email{mcschroder@wisc.edu}
\affiliation{
  \institution{
  Department of Computer Sciences\\
  University of Wisconsin--Madison}
  \city{Madison}
  \state{Wisconsin}
  \country{USA}
}

\author{Yuna Hwang}
\orcid{0000-0001-7726-8003}
\email{yunahwang@cs.wisc.edu}
\affiliation{
  \institution{
  Department of Computer Sciences\\
  University of Wisconsin--Madison}
  \city{Madison}
  \state{Wisconsin}
  \country{USA}
}

\author{Callie Y. Kim}
\orcid{0009-0001-4195-8317}
\email{cykim6@cs.wisc.edu}
\affiliation{
  \institution{
  Department of Computer Sciences\\
  University of Wisconsin--Madison}
  \city{Madison}
  \state{Wisconsin}
  \country{USA}
}

\author{Leqian Cheng}
\orcid{0009-0007-0320-2010}
\email{lcheng89@wisc.edu}
\affiliation{
  \institution{
  Department of Computer Sciences\\
  University of Wisconsin--Madison}
  \city{Madison}
  \state{Wisconsin}
  \country{USA}
}

\author{Jeffrey Li-cheng Liu}
\orcid{0009-0002-0033-2139}
\email{jlliu3@wisc.edu}
\affiliation{
  \institution{
  Department of Computer Sciences\\
  University of Wisconsin--Madison}
  \city{Madison}
  \state{Wisconsin}
  \country{USA}
}

\author{Chenchen Zheng}
\orcid{0009-0008-4191-7241}
\email{czheng88@wisc.edu}
\affiliation{
  \institution{
  Department of Computer Sciences\\
  University of Wisconsin--Madison}
  \city{Madison}
  \state{Wisconsin}
  \country{USA}
}

\author{Xinning He}
\orcid{0009-0009-4869-1051}
\email{xinning.he@wisc.edu}
\affiliation{
  \institution{
  Department of Computer Sciences\\
  University of Wisconsin--Madison}
  \city{Madison}
  \state{Wisconsin}
  \country{USA}
}

\author{Bilge Mutlu}
\orcid{0000-0002-9456-1495}
\email{bilge@cs.wisc.edu}
\affiliation{
  \institution{
  Department of Computer Sciences\\
  University of Wisconsin--Madison}
  \city{Madison}
  \state{Wisconsin}
  \country{USA}
}

\renewcommand{\shortauthors}{Schroder et al.}

\begin{abstract}
As collaborative robots have entered domains such as manufacturing, agriculture, and healthcare, programming or adapting robot behavior typically requires robotic expertise that most end users lack. Natural language lowers this barrier. Recent advancements in large language models (LLMs) have made it feasible to translate natural language into robot task plans. However, language-based task specification can suffer from semantic ambiguity, and generative models lack transparency for how language instructions are translated into robot actions, making it difficult for users to validate plans before execution. To address these issues, we introduce \system: \texttt{\textbf{S}imulation-driven \textbf{H}uman-in-the-loop \textbf{R}efinement \textbf{I}nterface for \textbf{M}anipulation \textbf{P}lanning}. \system\ allows users to automatically generate a hierarchical robot primitive plan using natural language and iteratively revise their plan through re-prompting and explicit correction. At each revision, \system\ allows users to validate their plan in simulation, and once satisfied, execute it on the physical robot. Through a user study involving participants planning tabletop kitchen tasks ($N=35$), we validate that \system\ improves perceived control and enhances robot transparency. \revise{}{System videos and source code are available at} \href{https://wisc-hci.github.io/SHRIMP}{https://wisc-hci.github.io/SHRIMP}.

\end{abstract}



\begin{teaserfigure}
  \centering
  \includegraphics[width=\textwidth]{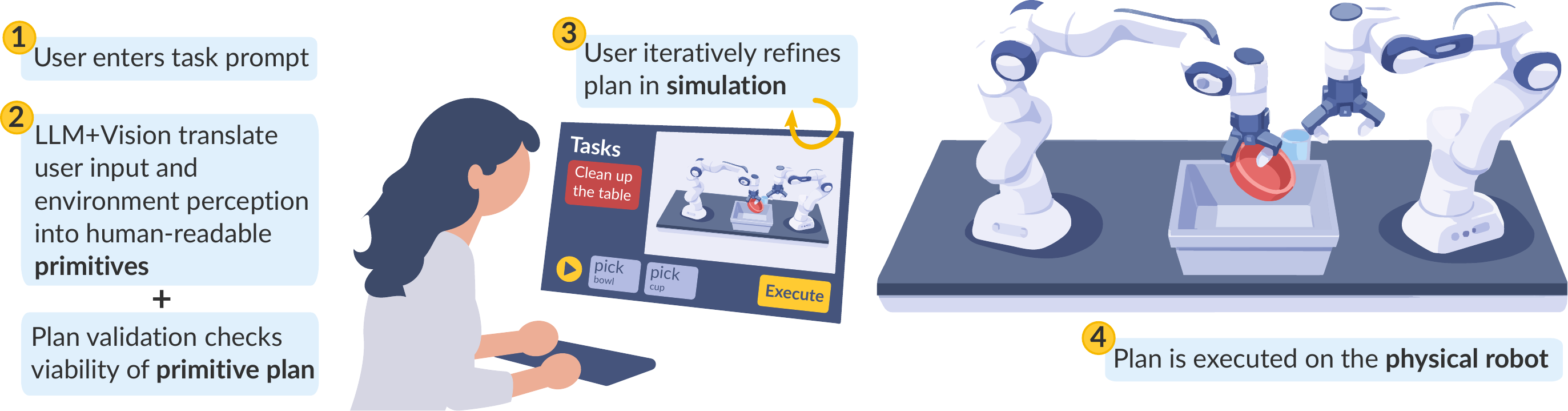}
  \caption{We introduce \system, a system that allows users to automatically generate a hierarchical robot primitive plan using natural language and iteratively revise their plan in simulation through re-prompting and explicit correction, before executing it on the robot.
  }
  \label{fig:teaser}
\end{teaserfigure}


\maketitle

\section{Introduction}

% Why automated end-user robot planning is important
Collaborative robots (\textit{i.e.}, cobots) have been deployed to manufacturing, automotive, agriculture, and healthcare contexts \cite{Taesi-2023, Pietrantoni-2024}. These robots are designed to operate alongside humans to improve productivity and safety and to be reprogrammed on-the-fly by their users to adapt to various tasks \cite{Taesi-2023}. As these users may not always have robotics expertise, successful integration requires user-friendly task planning interfaces that interpret high-level, operator intent, such as a robot skill or task \cite{Taesi-2023, Pietrantoni-2024}. Current non-expert cobot programming tools, such as \revise{the}{a} teach pendant, remain difficult for non-experts to use \cite{Dong-2021, Giannopoulou-2021}, as they operate at a low-level, requiring users to specify robot joint positions and construct plans step-by-step from scratch. These limitation\revise{}{s} motivate the development of alternative interface modalities that support higher-level robot planning. 

% Issues with current generative AI (LLM, VLA)
Natural language offers an accessible way for non-expert users to convey intent for robot task plans. Recent advances in Large Language Models (LLMs) and Vision-Language-Action (VLA) models now make it feasible to translate this input into executable plans. These models have demonstrated capabilities in reasoning and semantically-valid decision making \cite{Wang-2025, Zitkovich-2023}. However, a key limitation of language-based task specification is its inherent imprecision and ambiguity (\textit{e.g.},  the statement, ``move the object left,'' does not specify the movement distance, whether the reference is ego- or exo-centric, or which object should be manipulated) \cite{8298518, 11247661, ren2023robotsaskhelpuncertainty}. This limitation is especially problematic in robot planning, where tasks must be translated into precise pose goals and trajectories. Additionally, generative models, especially VLA models, are often ``black-boxes,'' limiting the users' ability to validate task plans before execution on the robot. Even when LLMs produce plans in textual format, or provide user-interpretable explanations, complex robot plans can still be difficult for the user to interpret and verify before execution \cite{Chen_Huang_2024}.

% Summary of our interface
To address the ambiguity of language and the limited transparency of generative models, while keeping the accessibility of language for non-experts, we propose \system: \texttt{\textbf{S}imulation-driven \textbf{H}uman-in-the-loop \textbf{R}efinement \textbf{I}nterface for \textbf{M}anipulation \textbf{P}lanning}. \revise{\system{} allows users to specify tasks in natural language, automatically generates a corresponding robot plan using an LLM, and supports iterative review and refinement of the plan.}{\system{} allows users to specify tasks in natural language, automatically generates a corresponding robot plan using an LLM, and represents that plan as a hierarchy of primitives that can be reviewed and refined in simulation. This combination of structured plan decomposition and physics-grounded simulation supports a  complete iterative loop of inspection, correction, and verification.} \revise{}{Unlike existing LLM-based robot planning systems} \cite{Ge-2024, karli-2024}\revise{}{, \system{} does not simplify perception and does not rely heavily on text- or code-centric verification and correction.}

\system\ prompts users to disambiguate their task within an iterative plan refinement process, where they adjust the sequence of primitives in a plan or the parameter values of those primitives (\textit{e.g.}, robot end-effector poses) via natural language re-prompting or explicit correction. After each update, users can test their plans in simulation, to ensure that they align with their original intent before executing on the physical robot. The simulation supports full-execution and step-by-step primitive action playback, and acts as a safe environment to facilitate iterative refinement without the risk of physical collisions or damage.

\system\ improves plan transparency by presenting each plan as a sequence of high- and low-level robot primitives. The primitive representation gives users control over the level of abstraction at which they wish to engage with the plan. Additionally, the simulation provides transparency into the expected execution of the plan. Because our simulation is fully physics-based, it shows realistic object manipulations, allowing users to identify issues such as unintended object motion or incorrect placement that are not implemented in prior work that solely rely on trajectory-only visualizations \cite{Darboven-2024, karli-2024}.

To evaluate \system, we perform an ablation study ($N=35$), where we justify the importance of our system features. Our study shows that \system\ provides users with increased control and improved transparency compared to the baseline, in which no robot primitive plans are visible nor editable. 

Our work makes the following contributions:
\begin{enumerate}[topsep=0pt]
    \item \textit{System Contribution}: \revise{\system{} enables transparent, iterative refinement of natural language-based robot plans through task breakdown and primitive playback in simulation.}{\system{} enables physics-grounded robot plan iteration by allowing users to refine generated plans through multi-level language and primitive editing, increasing transparency and control  during refinement.}
    \item \textit{Empirical Contribution}: We evaluate the effectiveness of \system\ through an ablation study and investigate user-perceived levels of control and transparency over LLM-generated robot plans. 
    \item \textit{Design \& Research Implications}: We present design and research implications on robot planning based on our evaluation. We provide insights on how to improve end-user transparency and control.
\end{enumerate}

% This is placed here so that it is at top of page before Interface description
\begin{figure*}[t]
    \includegraphics[width=0.9\textwidth]{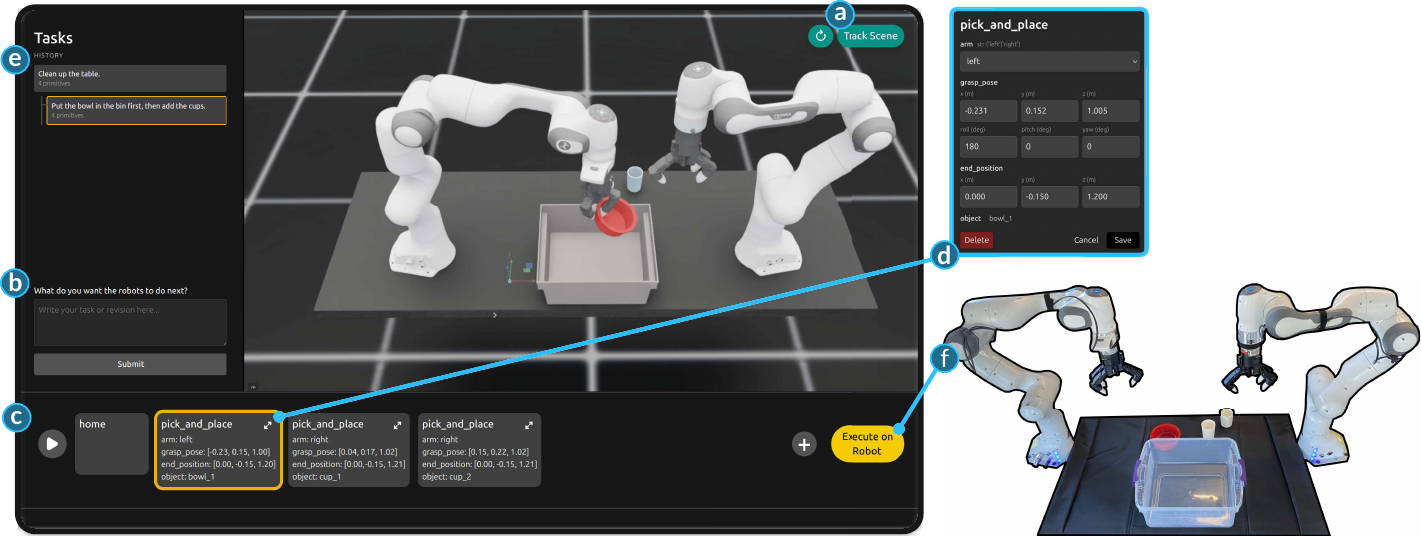}
  \caption{ \textit{\system\ Interface} --- This diagram outlines the user workflow with the interface, from submitting a natural language task prompt to executing the plan on the physical robot. A description of steps a--f is provided in Section~\ref{user-workflow}.}
  \label{fig:UI}
\end{figure*}

\section{Related Works}
\subsection{End-User Robot Task Planning Systems}
With the increasing prevalence of robots in various domains, there is much research done on end-user robot programming (EURP). End-user robot programming, by definition, aims to keep non-robot experts in-the-loop for the creation and maintenance of robot programs (see \cite{Ajaykumar-2021} for a survey of existing EURP systems). 

To alleviate user burden while programming robotic systems, many programming modalities have been implemented and tested, including interfaces that rely on visual programming paradigms, augmented and mixed reality (AR/VR) devices, programming-by-demonstration (PbD) techniques, natural language programming paradigms, and tangible interfaces \cite{Ajaykumar-2021, 10.1145/3472749.3474773, 10.1145/3332165.3347902, pais2015learning, 11246326, 10974232, 10.1145/3332165.3347957, 10.1145/3242587.3242634, 10.1145/3643834.3660721}. Our work connects to existing EURP interfaces that are visual-based and those that operate based on natural language. In order to further assist end-users in communicating their programming intent, many previous works use the concept of primitives, which are units of robot actions that are exposed to the user to imply feasible, robot functionalities. 

For visual programming interface, block-based programming interfaces are prevalent \cite{10.1145/2909824.3020215, 10.1145/3610978.3640644, 10.1145/3610977.3637477, 10.1145/3610977.3634974, Alexandrova-2015, Beschi-2019}. Other EURP systems provide users with more granular and hierarchical structures of robot programs, where \citeauthor{Guerin-2015} \citep{Guerin-2015} express robot program control flows as block-based, tree structures, and \citeauthor{Liang-2019} \citep{Liang-2019} distinguish between high- and low-level robot primitives and display it on a visual interface, which is the closest to our work.

Additional existing works on robot task specification take advantage of natural language programming. There have been many works in this realm even prior to the advent of LLMs, where users could easily type in their instructions in a text box. For example, \citeauthor{Beschi-2019} \citep{Beschi-2019} and \citeauthor{10.1145/2909824.3020215} \citep{10.1145/2909824.3020215} implemented a natural language interaction component into their system, where users can enter the desired robot behaviors and the system parses those into pre-defined robot actions. However, these systems are rigid due to hard-coded features and no support for revising instructions.

\subsection{Robot Task Planning with LLMs}
Robotic task planning interfaces have increasingly leveraged LLMs to bridge the gap between high-level human instructions and low-level robot execution \cite{10500490, 10.1145/3757279.3785550}. 
Early works focused on grounding LLM reasoning in physical affordances to ensure that generated plans are executable. Systems like SayCan \cite{ahn2022icanisay} combine LLM-generated actions with learned value functions to select feasible behaviors, while VoxPoser \cite{huang2023voxposercomposable3dvalue} constructs composable 3D value maps from language to guide motion planning in open-vocabulary settings. Vision-Language-Action (VLA) models such as RT-2 \cite{Zitkovich-2023} enable end-to-end mapping from semantic inputs to robot actions through large-scale co-training. Other work, like ProgPrompt \cite{Singh-2023} represents tasks as executable code with assertions and recovery steps. AutoGPT+P \cite{Birr-2024} combines LLMs with classical planners by translating language into symbolic goals and adapting to missing objects. However, such systems typically \revise{}{lack} interfaces for users to review or modify generated plans, making them difficult to verify and limit their reusability across tasks.

To address this limitation, recent work has begun to explore human-centric interfaces that aim to make robot programming more accessible. Systems such as Alchemist \cite{karli-2024} and Cocobo \cite{Ge-2024} enable users to author and modify robot behaviors through natural language, code, or graphical abstractions, and \citet{Dong-2025} incorporates retrieval-based knowledge to support domain-specific programming. \citet{Merlo-2025} allows the use of natural language feedback to assist with revisions during plan execution; however, in their system, revision is limited to verbal communication. Also, these systems remain limited in both capability and usability: perception is often constrained to simplified settings and interfaces are largely text- or code-centric. As a result, users must rely on manual inspection to verify generated plans, with little support for interactive editing or simulation prior to deployment.
This highlights the need for systems that tightly integrate grounded perception, structured planning, and intuitive interfaces, enabling users to interactively inspect, refine, and validate robot task plans before execution.

% This is placed here so it ends up at top of page in below the user interface section
\begin{figure*}[t]
    \includegraphics[width=\textwidth]{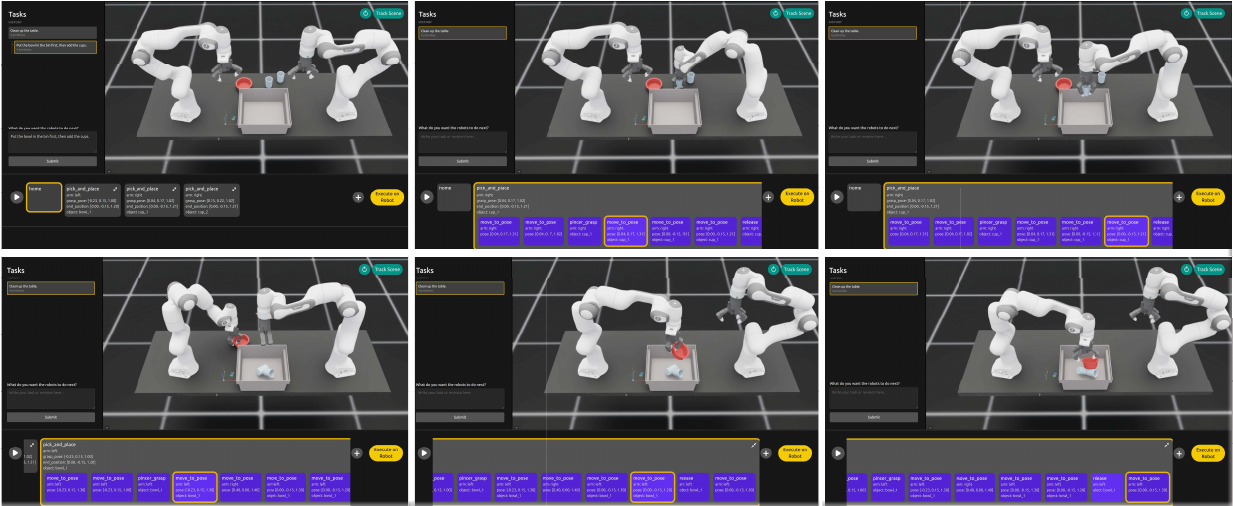}
  \caption{ \textit{Plan Interaction} --- After the initial simulation execution, users can step through the plan by selecting each primitive (yellow highlight), at either the high level (gray primitives) or low level (purple primitives), to observe the robot executing each step. In this particular example, the task is to \tThree.}
  \label{fig:timelapse}
\end{figure*}

\subsection{Robot Plan Visualization \& Simulation}
Previous research has examined many different forms of end-user-visible robot plan representations.  \citet{10974179}, for example, provides users with opportunities to manipulate the environment to build a timeline of visual plan steps \cite{10974179} and \citet{Darboven-2024} use RViz to visualize robot plans so that end users can detect errors before deployment \cite{Darboven-2024}. Overall, these examples show the importance of visualizing robot behavior prior to task execution; however, they primarily focus on either previewing robot motion or inspecting the robot's plan, rather than iteratively planning robot manipulation based on the richness of the robot's interactions with objects in its environment.

More recently, user interfaces have emerged that support plan revision. In particular \citet{Kuts-2022} presents a simulation-based digital twin planning environment that uses game-engine physics to conduct realistic manipulation simulations. Similarly, \citet{Matthaiakis-2017} uses simulation to allow users to reference and modify robot plans \cite{Matthaiakis-2017}. However, these systems do not take advantage of automatic plan generation, nor do they include general perception of the current scene. Our work extends these research directions by integrating automated planning with a physics-based interactive simulation to provide an avenue for iterative plan improvement.

\section{The \system\ Interface}

\subsection{User Workflow}
\label{user-workflow}
\system\ supports end-user creation of tabletop task plans for a bi-manual robot arm system. The interface contains a digital twin of \revise{the}{} the robot setup and objects so the plan can be tested in simulation before being executed on the robot  (see Figure~\ref{fig:UI}). This encourages users to continuously edit and revise their plan in a safe environment. Additionally, since the simulation is physics based, users can verify object-robot interaction within the plan (\textit{i.e.}, if a bowl is dropped at the edge of the bin, does it land in the bin or fall off the side). The user interaction flow is illustrated in Figure~\ref{fig:UI} and described as follows. 
\\ \raisebox{-0.25\height}{\includegraphics[width=0.4cm]{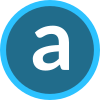}} Users track the scene to replicate objects from the physical environment within the simulation. 
\\ \raisebox{-0.25\height}{\includegraphics[width=0.4cm]{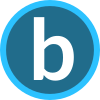}} Then, users enter their task prompt (\textit{i.e.},  ``Clean up the table by putting everything in the box'').
\\ \raisebox{-0.25\height}{\includegraphics[width=0.4cm]{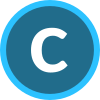}} The hierarchical primitive plan is automatically generated, which users can execute in simulation. The plan consists of both high- and low-level robot primitives. Users can expand the high-level primitives to view the low-level primitives. After initial execution, users can step through the hierarchical plan (at either the high- or low- level) to observe what happens after each primitive (see Figure~\ref{fig:timelapse}).
\\ \raisebox{-0.25\height}{\includegraphics[width=0.4cm]{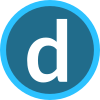}} Users can edit the plan by either re-prompting, re-ordering/adding primitives, or modifying the primitive parameters. While pose parameters are being edited, a red arrow appears in simulation to indicate the pose. 
\\ \raisebox{-0.25\height}{\includegraphics[width=0.4cm]{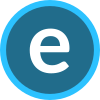}}  Any new or edited plans appear in Task History, under the parent plan. Users can revert back to previous plans if desired and iterate from there.
\\ \raisebox{-0.25\height}{\includegraphics[width=0.4cm]{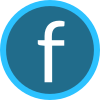}} Once satisfied with their plan, users can execute it on the physical robot.

% These are placed here so they end up in the proper section
\input{tables/primitives}

\begin{figure*}
    \includegraphics[width=0.85\textwidth]{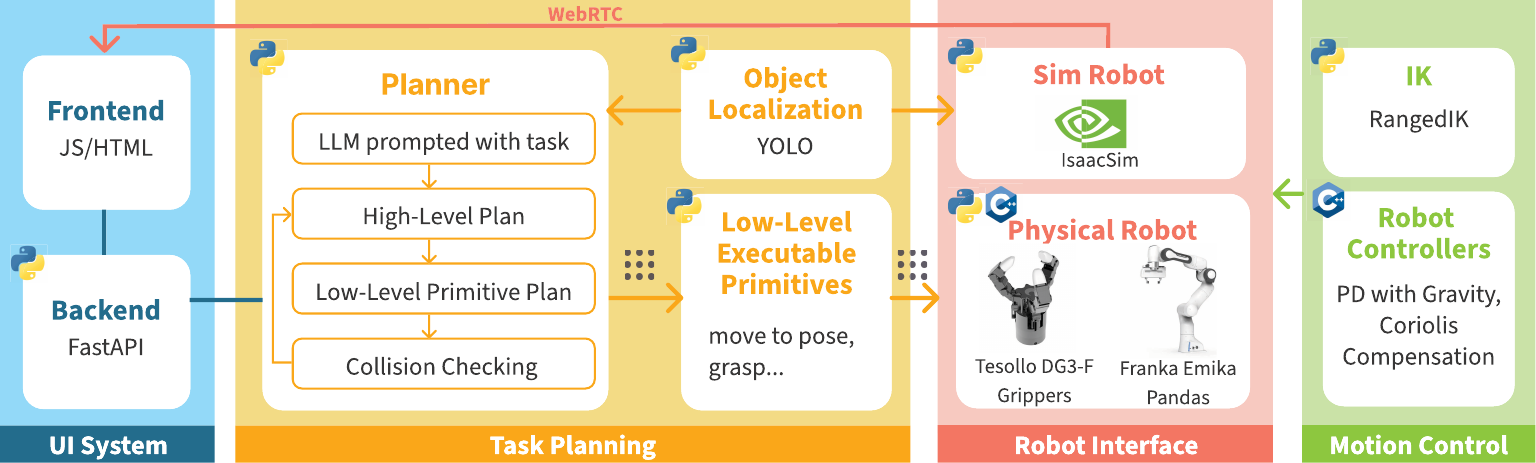}
  \caption{{System Architecture} --- Our hardware and software architecture is shown in this diagram.}
  \label{fig:architecture}
\end{figure*}

\subsection{Plan Generation}
\subsubsection{Primitives}
The hierarchical robot plan consists of low- and high-level primitives. Low-level primitives are the basic building blocks of robot actions. High-level primitives are sequences of low-level primitives. Each primitive has its own parameters.\revise{}{ This hierarchical primitive decomposition is an explicit design choice that enables step-level plan inspection, simulation, and parameter editing.} See Table~\ref{tab:primitives} for the full list of primitives that we developed.

\subsubsection{LLM Generation}
\label{sec:llm_generation}
 We used GPT 5.1 \cite{Singh-2025} as our LLM\revise{}{, selected for its accessibility, reproducibility, and reliability with our system}. The LLM is provided with the robot configuration, list of available primitives, the positions of all the objects in the scene, and the user's natural language task description. The LLM is prompted to generate the sequence of primitives needed to complete the plan, prioritizing high-level primitives.  See Appendix~\ref{sec:planExamples} for examples of participant task prompts and resulting plans. 

\subsubsection{Plan Verification}
\label{sec:plan_verification}
After the plan is generated, it undergoes three validation checks to ensure robustness. The first check confirms that grasp primitives with object parameters correctly target the intended object. This is done by calculating expected object and robot positions at each step of the plan and checking that the prior move\_to\_pose primitives put the gripper within grasping range of the object. If the primitive parameters position the robot gripper too far from the object, the pose parameters are replaced with those from the object tracking. The second check is to prevent arm-to-arm collision. If both robots' end-effectors are too close to each other, we insert retract motions into the plan to prevent collisions. The final check is to prevent robot-object collision. If we detect that the arm (or the object the arm is holding) collides with other objects in the scene during transit, we increase the height of the robot's trajectory to prevent collisions.

\subsection{Hardware \& Software Architecture}

Figure~\ref{fig:architecture} shows our architectural diagram. Below we describe the components, progressing from the hardware robot setup to the frontend interface.

The system consists of two 7-degree of freedom Franka Emika Panda Robot arms, each mounted with a 12-DOF Tesollo 3 Finger gripper. The simulation engine is Isaac Sim. Both the physical robots and the simulated robots share the same motion control layer so that the simulated robot acts as a digital twin of the real robot. For  IK (inverse kinematics), we use RangedIK \cite{Wang-2023}. For both the simulated and physical real Franka robots, we developed a PD (Proportion-Derivative) Torque controller with Coriolis compensation. Since the physical Franka Control Interface (FCI) automatically performs gravity compensation, we add additional gravity compensation to the simulation controller. The simulated Tesollo grippers use the same controller; however, for the physical Tesollo grippers, torque is approximated using duty cycle, as they do not support direct torque control. Note that the simulation (and  therefore the control loop) runs in real-time at approximately 200 Hz, compared to the Franka that run at 1000 Hz and the Tesollo at 500 Hz. To compensate for the lower control rate, the simulation controller is tuned with higher $K_p$ gains to match the physical robots' motion response.

We use a single RGB-Depth d435 Realsense Camera to \revise{to}{} localize objects in the scene using YOLO11 \cite{khanam2024yolov11overviewkeyarchitectural}. This object location data is used by both the simulation and planner. The planner receives the task prompt from the UI backend, creates a plan, and breaks it down into low-level primitives which are executed using the robot interfaces. The plan breakdown and verification steps are described in sections \ref{sec:llm_generation} and \ref{sec:plan_verification}.

Our backend is a REST API implemented using FastAPI \cite{Ramirez-2018}. User plans are stored in file storage as JSON. Our frontend is built using pure Javascript and HTML, with Tailwind \cite{Wathan-2017} as our CSS library. We implemented our own lightweight state storage based on the observer pattern. The simulation is streamed to the frontend using WebRTC \cite{Nvidia-2023}.

To run this entire system, we use two computers (Figure~\ref{fig:system_setup}), one that has a \revise{Realtime}{Real-Time} Kernel Patch to run the control logic and interfaces for the physical robots; and another with \revise{a Nvidia}{an NVIDIA} GPU to run the simulation, along with the UI and Task Planning components. In order to communicate between these two desktops, we \revise{us}{use} ROS2 Jazzy.

\section{\system\ Evaluation}

\begin{figure}
  \centering
  \includegraphics[width=\columnwidth]{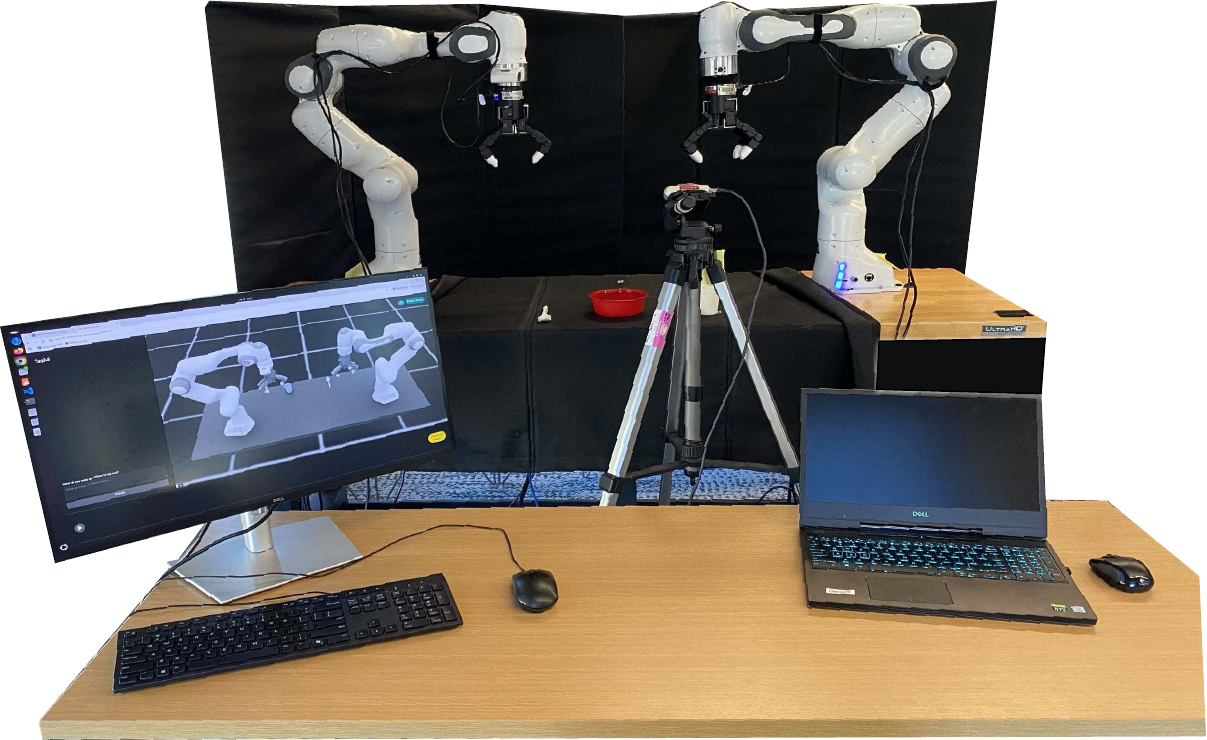}
  \caption{ \textit{System Setup} --- The robot hardware consists of two Franka Emika Panda Robot arms, each mounted with a Tesollo 3D-GF gripper, positioned on a table to perform tabletop tasks. A RGB-Depth d435 Realsense Camera is positioned to track the scene. The system runs on a Desktop with a \revise{Nvidia}{NVIDIA} GPU (left) and a laptop  with a \revise{Realtime}{Real-Time} Kernel Patch (right).}
  \label{fig:system_setup}
\end{figure}

We ran an ablation within-participants experiment to understand how users interact with the interface  ($N=35$; ages 18---65+; 14 male, 19 female, 2 preferred not to answer). Each study lasted approximately 50 minutes, for which participants were compensated \$15 USD. The study was conducted in front of the two bimanual arms with a desktop for the participant to use the interface, and a laptop for the participant to watch the video tutorials and take the survey (see Figure~\ref{fig:system_setup}).

\subsection{Study Design}
Participants were given one of three tasks: \tOne, \tTwo, and \tThree\ (described in Table~\ref{tab:tasks}). For their given task, participants performed three trials, using three different versions of the interface. The three ablated versions of the interface are described below.
\\ \cThree: In this condition, users use the full system, with all features enabled, as described in Section~\ref{user-workflow}.
\\ \cTwo: In this condition, users are unable to edit primitive parameters, expand, or step through the hierarchical plan. Only the high-level primitives are visible, along with arm and object parameters. \revise{}{This condition was designed to test whether exposing a high-level plan alone is sufficient to support users, without introducing lower-level inspection and editing affordances. This is similar to existing LLM and VLA systems that expose only high-level plans and provide limited access to underlying actions.}
\\ \cOne: In the baseline condition, users are unable to view the primitive plan. \revise{This is similar to}{This condition reflects} existing LLM and VLA \revise{approaches, where users cannot see nor interact with the plan, and can only re-prompt to change the plan.}{interaction approaches that treat the planning process as a black box, allowing users to revise robot behavior only through natural-language prompts.}

The order of conditions was counterbalanced across conditions to control for learning effects. 

\input{tables/tasks}

\subsection{Procedure}
The study was approved by the Institutional Review Board of \University. All participants signed a consent form prior to starting the study. The experimenter first gave the participants a brief introduction of the experiment and showed a video tutorial that outlined how to use the different features present in each condition. The experimenter then presented the task to be performed. Participants were given up to 10 minutes to complete planning for the task, after which they could opt to execute on the robot if they had not already done so. After the trial was finished, participants completed a post-trial evaluation for the interface condition that was just given. This process was then repeated for the remaining two conditions. After completing all trials, participants answered open-ended questions about the overall system and provided demographic information.

\subsection{Hypothesis}
We formulated three hypotheses: Participants with \cThree\ condition will have greater perceived control compared to \cTwo\ and \cOne\ (H1); Participants using the \cThree\ condition will feel the system is more transparent compared to \cTwo\ and \cOne\ (H2); Participants using condition \cThree\ will view the system as being more usable compared to \cTwo\ and \cOne\ (H3).

\subsection{Measures}
In order to validate the prior \revise{hypothesis}{hypotheses}, we collected the following set of data. While the participant was interacting with the interface, we logged all user input and feature usage. After each trial, the participant completed four questionnaires: a subset of the Transparency Of Robots Scale (TOROS) \cite{Angelopoulos-2025} ($\alpha=0.89$), the Perceived Autonomy Scale (with language changed to first-person) \cite{Sankaran-2021}  ($\alpha=0.87$),  the System Usability Scale (SUS) \cite{Brooke-1996} ($\alpha=0.90$), and the TASK Load Index (NASA-TLX) \cite{Hart-1998} ($\alpha=0.78$). We also recorded trial completion time and binary success of each of the three subtasks (see Table~\ref{tab:tasks}), summed together to obtain a total task success score from 0 to 3. At the end of all three trials, we also asked a set of general open-ended questions and demographic questions. Appendix~\ref{sec:survey_questions} contains the questionnaires and open-ended questions.

\subsection{Analysis}
The Perceived Autonomy, TOROS, NASA-TLX, and SUS score measures, along with the success and time measures were analyzed using linear mixed effects models (LMMs) with the \textit{statsmodels} Python library \cite{Seabold-2010}. We report comparisons from three models: one with the condition as a fixed effect, one with the task as a fixed effect, and the other with the condition, task, and their interaction as fixed effects. \revise{Both}{All} models include participants as a random effect to account for the within-subjects design. We also compute Pearson correlation between measures, and Spearman correlation between trial progression and measures using the \textit{scipy} Python library \cite{Virtanen-2020}. For each scale measures, we also computed Cronbach's Alpha using the \textit{pingouin} Python library \cite{Vallat-2018}.
 
For the qualitative data, we conducted a Thematic Analysis \cite{braun-2006} of the user language prompts (recorded from the interface logging), as well as the open-ended questions. One researcher coded the data and created a codebook. Using the codebook, a second researcher coded the data again. The codebook was then revised and data re-coded iteratively until a\revise{}{n} Inter-Rater-Reliability percent agreement  of $\geq 90\%$ was reached.

% Placed here so it ends up in the correct section visually
\begin{figure*}[t]
  \includegraphics[width=\textwidth]{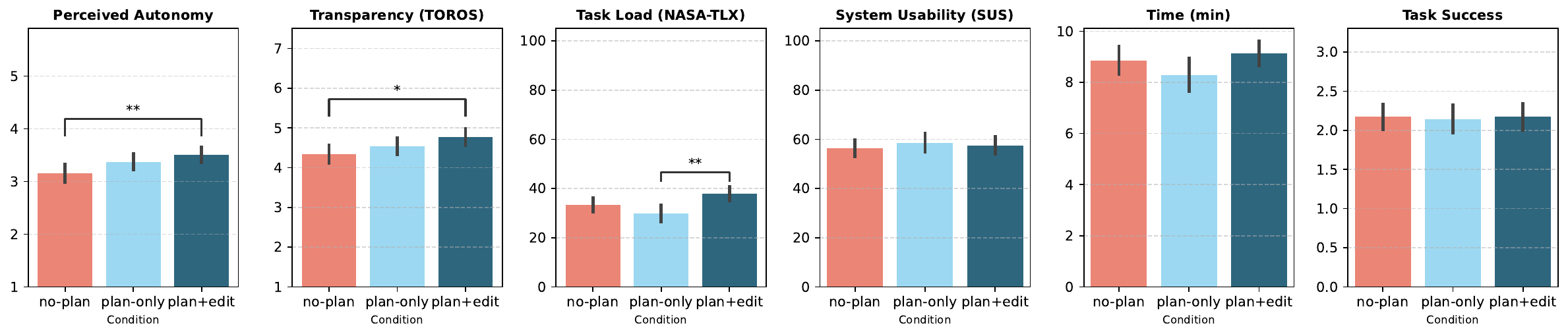}
  \caption{ \textit{Quantitative Measures Across Conditions} --- The bar chart shows the mean participant scores for  Perceived Autonomy (control), Transparency, Task Load, System Usability, and mean measures for Time (minutes) and Task Success (out of 3). For Perceived Autonomy, Transparency, System Usability, and Task Success a higher value is better, while for Task Load and Time a lower value is better. Horizontal brackets indicate significant pairwise comparison $(p < 0.05^{*},\ p < 0.01^{**})$. Vertical lines on each bar indicate standard error.}
  \label{fig:mean_full}
\end{figure*}

\section{Results}
In our analysis, we explore how viewing and interacting with the hierarchical primitive plan impacts the users' experience with the system. Figures ~\ref{fig:mean_full} and ~\ref{fig:mean_by_task} summarize our results, with additional visualizations provided in Appendix~\ref{sec:extra_figures}.
Overall, participants reported significantly higher perceived autonomy ($p=0.007$) for \cThree\ compared to \cOne\ (\cOne: $M=3.15, SD=0.99$; \cThree: $M=3.50, SD=0.84$). % \cTwo: $M=3.37, SD=0.90$; 
Participants also reported significantly higher transparency (TOROS) ($p=0.026$) for \cThree\ compared to \cOne\ (\cOne: $M=4.34, SD=1.34$; \cThree: $M=4.77, SD=1.27$). % \cTwo:$M=4.54, SD=1.25$; 
There were no significant results between conditions \cTwo\ vs \cOne\, nor \cTwo\ vs \cThree\ for perceived autonomy and transparency. This suggests that viewing the \revise{}{high-level} plan is not sufficient. Instead, users need both the viewing and interaction features of \cThree\ for higher autonomy and transparency.

Although participants rated the \cThree\ system to be the most transparent and provide them with more control, participants also reported significantly higher perceived task load (NASA-TLX) ($p=0.002$) for \cThree\  compared to \cTwo\ (\cTwo: $M=29.92$, $SD=20.24$; \cThree: $M=37.86$, $SD=16.30$).
% (\cOne: $M=33.41$, $SD=16.82$; \cTwo: $M=29.92$, $SD=20.24$; \cThree: $M=37.86$, $SD=16.30$).  
There were no significant differences between conditions for the usability rating (SUS) or task success.
% (\cOne: $M=56.28, SD=20.39$; \cTwo: $M=58.57, SD=22.14$; \cThree: $M=57.5, SD=20.53$), time in minutes (\cOne: $M=8.86, SD=3.23$; \cTwo: $M=8.28, SD=3.82$; \cThree: $M=9.14, SD=2.84$), or task success (\cOne: $M=2.17, SD=0.95$; \cTwo: $M=2.14, SD=1.06$; \cThree: $M=2.17, SD=0.98$). 

Below, we group our significant quantitative and qualitative findings into four themes.

\begin{figure}[b]
  \includegraphics[width=\columnwidth]{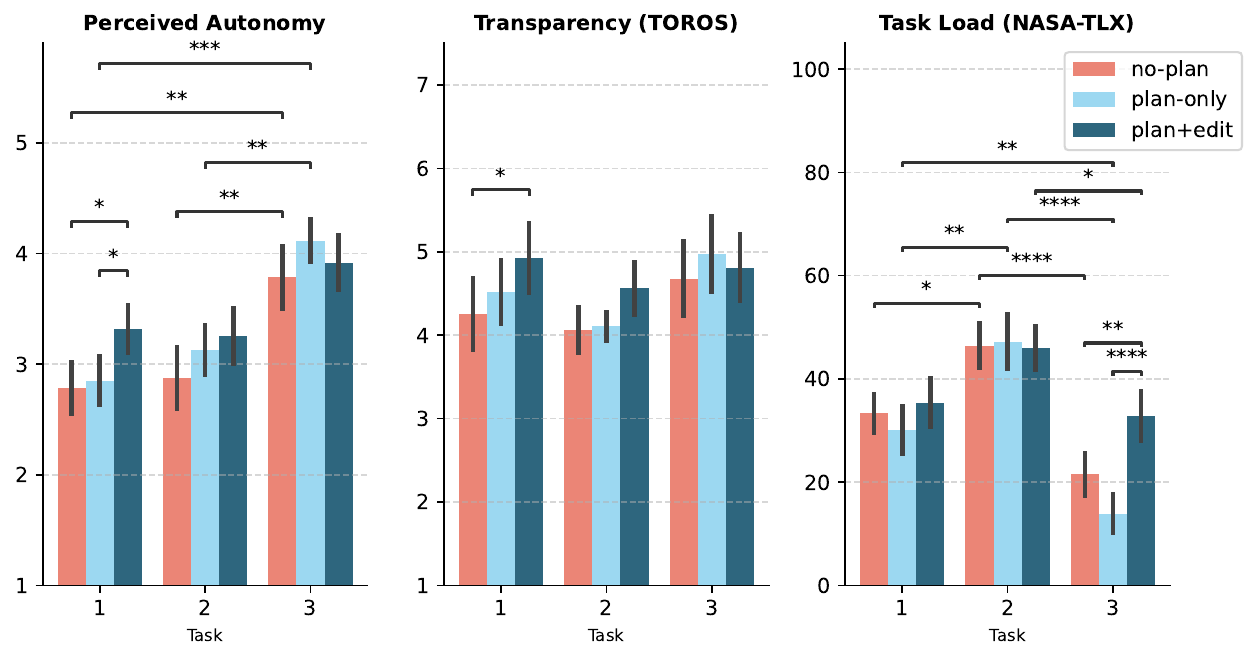}
  \caption{\textit{User Experience Measures Across Tasks} --- The bar chart shows the mean participant scores per task for Perceived Autonomy (control), Transparency, and Task Load. Task 1 is \tOne, 2 is \tTwo, and 3 is \tThree. Horizontal brackets indicate significant pairwise comparison $(p < 0.05^{*},\ p < 0.01^{**},\ p < 0.001^{***})$. Vertical lines on each bar indicate standard error. }
  \label{fig:mean_by_task}
\end{figure}

% Placed here so it ends up visually in this section
\begin{figure*}[b]
  \center
  \includegraphics[width=\textwidth]{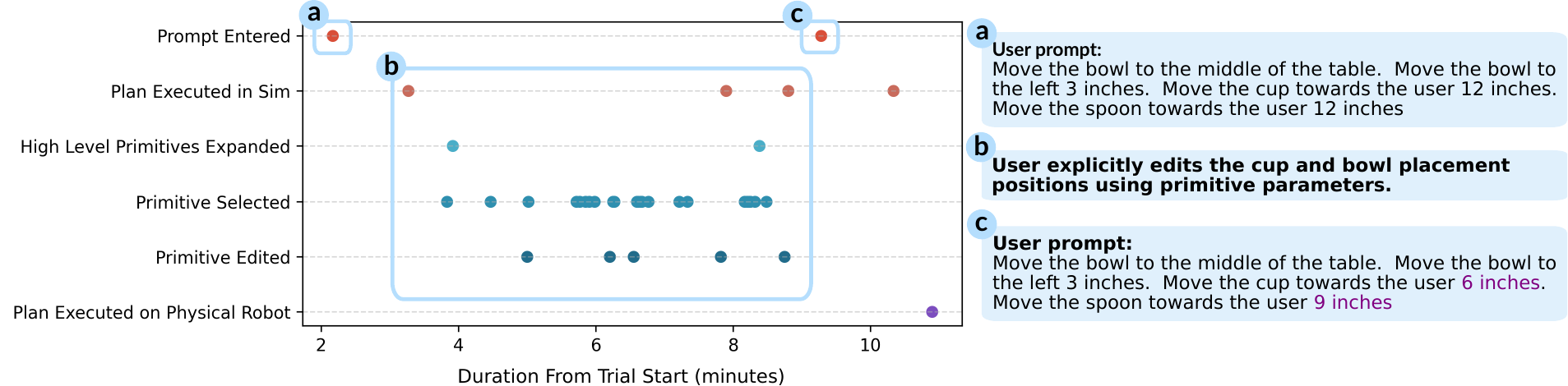}
  \caption{\textit{Event Timeline} --- Here we plot the time that a particular interface interaction event occurred during P29's \cThree\ condition.}
  \label{fig:events}
\end{figure*}

\subsection{Explicit Engagement with the Hierarchical Primitives Increases Perceived Control}
\label{sec:results-perceived-control}

Participants' perceived autonomy scores for the \cThree\ system were significantly greater than the \cOne\ baseline ($p=0.007$). This indicates that participants felt they had more control over the interface when they could explicitly step-through and edit the full hierarchical primitive plan. 22 participants mentioned how the \cThree\ features provided them with more control over the plan. For example, P26 said, \textit{``I liked the first \textup{[\cThree]}. Being able to edit specific parts of the robot's movements and tasks made me feel like I had more control and could be as simple or complex as I wanted it to be.''}

From the qualitative data, we see two main explanations for this effect:
\textbf{(1) Editing plan parameters provided non-\revise{ambagious}{ambiguous}, precise control.} For example, P17 explained  \textit{``I preferred the third trial \textup{[\cThree]} where I could make more precise adjustments in the software.''} \textbf{(2) It was sometimes easier to adjust the plan manually, than to put it into words.} For example, P21 said \textit{``The third \textup{[\cThree]} was obviously the best. It is much easier to input a value of `where to put object' than it was to try and describe with words where I wanted it to go.''}.

Additionally, although we did not find differences in usability between conditions, we did find that the perceived autonomy scores were positively correlated with the usability scores (r=0.635, p<0.001). This is described by P40, who said, 
\textit{
    I enjoyed the last trial \textup{[\cThree]} the most because I could be specific about each command and get the robot to do exactly what I wanted. I found myself wanting to be done with the first two trials \textup{[\cOne, \cTwo]} while I was using them because they weren't responding how I wanted. I wanted to spend more time working with the third system because I had more control over it and it became more fun and engaging.
}

Finally, although the \cThree\ condition le\revise{a}{}d to increased perceived control, it came at the cost of higher task load compared to condition \cTwo\ ($p=0.002$). The qualitative data explains that this is due to the participants viewing the \cThree\ as more complex because of the added plan information and editing features. For example, P35 said, \textit{``The 3rd trial \textup{[\cThree]} seemed complex; even though I was used to the interface, I wanted to check on the robot's responsiveness at each step which caused me to slow down.''} Eight other participants similarly referenced condition \cThree\ as being more complex and requiring more time to learn how to use the robot primitives. 

However, there is more nuance to consider, as the increase in task load is not consistent across tasks.  Task \tThree\ was the easiest of the three tasks, as indicated by Task \tThree\ having lower task load than \tTwo\ ($p<0.001$) and \tOne\ ($p=0.043$), as well as higher task success than \tOne\ ($p=0.024$), across all conditions. Incidentally, task load cost between conditions was only significant for task \tThree, not the other tasks (see Figure~\ref{fig:mean_by_task}). This indicates that for easier tasks, the added features in \cThree\ introduce unnecessary complexity. This is supported by the qualitative data of participants that performed \tThree. For example, P13 said, \textit{``Trial 2 \textup{[\cTwo]} was best. Trial 1 \textup{[\cThree]} was ok, but seemed unnecessary as the tracking was good enough that the xy positions were already correct.''} In contrast, for the two more complex tasks, there were no significant differences in task load between conditions, suggesting that the additional capabilities of \cThree\ may be more appropriate in these contexts.

\subsection{Viewing the Primitive Plan Improves Overall System Understanding}
Our participants scored condition \cThree\ as having significantly more transparency compared to \cOne\ ($p=0.026$). This was supported in the qualitative data where 19 participants explicitly expressed that viewing the primitives in \cThree\ provided them with more understanding of the plan. For example, when describing why they preferred condition \cThree\, P27 said, \textit{``I could see the steps spelled out in the boxes at the bottom of the screen. It was clear that each task would be completed in turn.''} Other participants explicitly stated that not seeing the plan in condition \cOne\ made them feel confused. For example, P38 said \textit{``The first trial \textup{[\cOne]} lacks clear explanation and I feel confused about how to make the prompt.''} 

Additionally, participants expressed that viewing the plan helped increase prompting (not just plan) transparency in two ways:
\textbf{(1) Users used the primitive sequence to see how and why the actual plan differed from their intended task prompt.} For example, P8 says \textit{``The interface feature where I could see each individual part of the task and how the system quantified it was the most useful because I was able to see where the task I entered could be shown differently by the system.''} 
\textbf{(2) Seeing the primitive plan helped the users better review and revise their prompt.} As P38 said, \textit{``The last trial \textup{[\cOne]} I did not see the breakdown of each robot movement, and I liked having that there so I could use it to better describe a certain step.''}

\input{tables/iteration}

\subsection{Users Interact with the \system\ Iteratively}
\label{sec:iteraction}

Across all conditions, participants revised their plans multiple times (\cOne: $M=3.68$, $SD=1.61$; \cTwo: $M=3.80$, $SD=2.53$; \cThree: $M=6.88$, $SD=4.30$). Participants also simulated their plan multiple times (\cOne: $M=4.65$, $SD=2.37$; \cTwo: $M=4.31$, $SD=2.94$; \cThree: $M=7.61$, $SD=4.07$). Figure~\ref{fig:events} shows an example of the iteration one participant went through until they were satisfied with their plan.

Three participants explicitly referenced that the simulation helped them iterate more efficiently. For example, P4 says   \textit{``The simulation helped, not having to try every iteration on the robot to save time and make it more efficient.''}

Additionally, using participants' language prompting data, we identified three major prompt iteration strategies (Table~\ref{tab:iteration} shows examples of each of these). 21 participants used the \sOne\ strategy, in which they tested subtasks one at a time, and then combined into one final plan at the end. 10 participants used the \sTwo\ strategy, in which they started off with a subtask, and then added another subtask to the plan at each iteration. 28 participants used the \sThree\ strategy, in which they specified their entire plan first, and then iterated by adding or changing specific details. Eight participants even chose to use the \sThree\ strategy as a way to update parameter information such as specifying spatial information. For example, P29 used the following prompt, despite having access to parameter editing in condition \cThree: \textit{``Move the bowl to the middle of the table.  Move the bowl to the left 3 inches.  Move the cup towards the user 12 inches.  Move the spoon towards the user 12 inches.''}

\subsection{Inconsistent Outcomes Lead to Frustration}
\label{sec:inconsistancy}
Six participants reported that similar prompts produced different plans. This variability is due to the non-deterministic nature of LLMs, where the same or similar prompts can result in different output. Sometimes a prompt similar to a previously successful one resulted in a failure due to primitive reordering that led to awkward joint configurations, preventing the robot from reaching the desired pose. However, since participants were not aware of this underlying cause, the inconsistency resulted in frustration. For example, P31 says, \textit{``The overall experience was fun and frustrating at the same time. I am still confused on why I was successful the first trial but unsuccessful the other two despite giving the same prompts.''} This pattern\revise{s}{} suggests inconsistent failures undermine usability, which is reflected in the quantitative data, where usability is moderately positively correlated with task success ($r=0.429$, $p<0.001$).

Participants compensated for this inconsistency by relying on the transparency the primitive plan provided and using the iteration strategies described in Section~\ref{sec:iteraction} to adapt their prompt. For example, P40 says, \textit{``The system was slightly frustrating, but I had an overall good experience. While unpredictable at times, once I understood how each trial worked, I was able to adapt my plans better.''} Consistent with these observations, task success was moderately correlated with trial progression, indicating \revise{}{a learning effect, where }users became more successful as they gained experience with the system ($r=0.337, p<0.001$). This is supported qualitatively, as 22 of our participants mentioned that they felt more proficient and comfortable with the system as trials progressed. For example, P32 said, \textit{``The first trial I was very insecure/nervous because I did not know what I was doing and I had never seen the robot move before so it was hard to shoot for an end goal I didn't know. By the third trial, I started to learn what command words would help the robot complete the task.''}

\section{Discussion}
In summary, our results confirm our first (H1) and second (H2) hypotheses that the \cThree\ system improves transparency and control compared to the baseline, \cOne. However, since there are no significant differences in transparency and control between \cThree\ compared to \cTwo, nor \cTwo\ compared to \cOne\, we conclude that \revise{}{the combined value of} both viewing and interacting with the \revise{}{full hierarchical} plan is required to improve transparency and control. Our results show that \system\ supports transparency by allowing users to inspect the primitives to understand how the generated plan differs from their intended prompt in order to revise their prompt. This outcome suggests that the primitive plan feature helps users to reduce their own language ambiguity. Users refined their language prompts iteratively, using \sOne, \sTwo, or \sThree\ strategies (see Figure~\ref{tab:iteration}).  Additionally, \system\ enables control through the primitive plan, where our results show that editing parameters grants the user precise control, especially when it is difficult to express desired changes with natural language. 

Our results do not support our final hypothesis (H3) that the \cThree\ system significantly improves usability. In fact, all conditions received below-average usability scores. We interpret this as reflecting a general lack of familiarity with the interface and task by participants, who are non-expert users that have limited experience with robots and simulation. Based on both \system's strengths and limitations, we introduce the following design and research implications for language-based robot planning systems. 

\subsection{Design Implications}
\textit{\textbf{Design Implication 1: \revise{}{Language-based robot planning s}ystems should support diverse prompting and editing strategies for efficient iteration.}} 
While prior language-based robot planning interfaces often implicitly assume that users will use a single, complete prompt, our results show that users employ three distinct prompting strategies: \sOne, \sTwo, and \sThree. The \sOne\ and \sTwo\ strategies are used by users to gradually build their own mental model of how the system works before writing a complete prompt, whereas \sThree\ appears to be used more directly to clarify intent. \revise{}{We hypothesize that end users may adopt these unique strategies because robotic tasks can include both temporal sequences of actions (\textit{e.g.}, pour then stir) and separate, standalone actions (\textit{e.g.}, placing one cup at a time in a bin without ordering constraints). Therefore}, language-based robot planning interfaces should support all \revise{three}{} of these prompting strategies \revise{}{to support practical robot programming practices}.

\revise{Beyond prompting behavior, o}{O}ur results also show that users prefer to have control and visibility over both high-level prompt revision and low-level primitive parameter editing to reduce language ambiguity. \revise{}{This demonstrates that}\revise{This preference reinforces our motivation that}{} language as a standalone modality is insufficient for robot task planning and revision, and that \revise{robot planning interfaces should accommodate}{} diverse editing strategies \revise{}{should be provided to assist end users to reason} across multiple layers of plan abstraction.

\revise{}{However, our results indicate that defaulting to exposing all possible prompting and editing strategies is also not the answer. As described in} Section~\ref{sec:results-perceived-control}\revise{}{, the complexity of the task can be used to determine whether or if any levels of assistance and prompting options are needed.}

For these iteration and editing strategies to be practical\revise{,however}{},  users must be able to update and evaluate their plans quickly. In our system, the simulation interface is key to allowing this efficient iteration.

\textit{\textbf{Design Implication 2: \revise{}{In language-based robot planning systems, l}ow-level plan details and editing controls should be introduced in a manner that reduces cognitive load.}}
\revise{In our evaluation, w}{}\revise{}{W}e found that the added control and transparency enabled by primitive editing \revise{can}{} also significantly increase\revise{d the}{s} users' cognitive load. This suggests a trade-off: while low-level plan details and editing controls are valuable, they should be presented in a way that mitigates task load. \revise{}{This also suggests that robot planning systems should support robot programmers with varying experience levels, as those with more experience may prefer specifying low-level, physics-level parameters, while novice users may prefer to operate in a higher-level.} 

\revise{Additionally, our participants had higher success with the system and expressed greater confidence and comfort with the system as trials progressed, indicating there was a learning-curve effect at play. These two observations suggest that one solution for reducing cognitive load is to gradually disclose features as users gain familiarity with the system and as task complexity increases.}{Additionally, since language-based robot planning systems, like other interfaces and planning tools, have a learning effect, they should gradually disclose features so that users can gain familiarity with the system and the increasing task complexity.}

\subsection{Research Implications}
\textit{\textbf{Research Implication 1: \revise{}{Language-based robot planning s}ystems should actively guide users across \revise{}{multiple} abstraction layers.}}
The ability to revise plans at both language and parameter levels is a core affordance of \system, enabling users to exercise precise control over their plans. However, simply exposing high- and low-level layers may be insufficient to achieve high usability. First, our findings suggest that it can be difficult for users to know which abstraction layer they should edit. For example, some participants who struggled with spatial placement repeatedly re-prompted \revise{}{using language} instead of \revise{directly}{} editing parameters, which would have been more effective and efficient. Additionally, users who think in terms of high-level goals (\textit{i.e.},  ``move the cup next to the bowl.''), may not naturally map that intent onto discrete numeric parameter fields without \revise{}{a direct, one-to-one mapping between elements in the language prompts and the low-level parameter fields (\textit{e.g.}, they may not know ``next to'' can be also described in the y-axis position parameter values)}\revise{scaffolding}{}. These observations highlight opportunities for future research. \revise{}{First, researchers should} investigate \revise{}{natural mappings between the low-level, physical layer and the high-level, language-based, semantic layer. Then, researchers can use that mapping to create efficient} interaction techniques that \revise{}{better support the users across these} abstraction layers. Rather than solely providing the correct tools for editing, these techniques should support users in developing a working mental model of which plan layers to engage with. \revise{One potential future direction is to design interfaces that visually link a generated parameter back to the phrase in the prompt that produced it.}{}

\textit{\textbf{Research Implication 2: \revise{}{Language-based robot planning s}ystems should help users localize failures across the pipeline, not just provide mechanisms to fix them.}} 
Despite the prevalence of generative AI-based end-user systems, limited research has explored how to assist users in identifying causes of generative-AI robot plan failures when they occur. This is especially detrimental and hence calls for future work, as end-user robot programming is iterative in nature and closely tied to task success and safety. \revise{Systems should be designed so that they prevent failure propagation across programming iterations.}{More research should be conducted in identifying and localizing the causes for robot program failures across programming iterations} \cite{Chen-2025, Errica_2025, banerjee2019fault}. 
\revise{As our participants have echoed sentiments of frustration over inconsistent plan outputs from the LLM (see Section~\ref{sec:inconsistancy}), and as few related works suggest future research should closely investigate how to assist end users to effectively identify and separate portions of natural language prompts that lead to faulty or inconsistent outputs.}{} This localization of failures is important as it further impacts perceived levels of autonomy and control over generative AI-based systems, where reducing unpredictability over AI system outputs will only then allow users to effectively evaluate these systems' performances.

\subsection{Limitations \& Future Work}
Our system has several technological and study design limitation, that present opportunities for future work. 

\textit{Technological Limitations} --- Plan generation is slow, taking up to ~1.5 minutes, especially when user task prompts exceed 100 words. This step slows down the iteration process, which some participants noted. Additionally, our vision localization system is sensitive to lightning, and as our study was conducted by a large window, we had to progressively adjust the blinds throughout the study to ensure the objects were being detected. Finally, to ensure consistent object grasping during the study, we manually defined grasp points relative to each object’s centroid. In the future, this function could be made more generalizable by using a machine vision model to automatically determine the best grasp points on objects.

\textit{Study Design Limitations} --- Some participants reported feeling time pressure during the first trial as they were still becoming familiar with the system. Task success may have improved if participants had been given a tutorial period to explore the interface. Additionally, for the sake of time, this study focused on tasks involving relatively few objects and subtasks. To stress test the system, future work could evaluate \revise{this system on a single, more complex task that requires many more plan iterations to complete successfully.}{ \system{}  with more complex or long-horizon tasks that require substantially more plan iterations.}

\section{Conclusion}
In this work, we introduce \system, a system that allows users to create robot plans using natural language and iterate upon them, using primitive breakdown and simulation execution to verify their plan. Through a user study, we show that our interactive primitive plan feature improves user perceived control and transparency. Additionally we offer design and research implications for creating robot planning interfaces that support control and transparency, while reducing cognitive load and maintaining usability.

\begin{acks}
\revise{}{This work was supported in part by National Science Foundation awards 2330040, ``Center: NSF Engineering Research Center for Human AugmentatioN via Dexterity (HAND),'' and 2152163, ``Integrating Robots into the Future of Work.''}
\end{acks}

\balance
\bibliographystyle{ACM-Reference-Format}
\bibliography{base, software}

\clearpage
\appendix

\section{Appendix: Questionnaires}
\label{sec:survey_questions}

\subsection{Perceived Autonomy Scale}
Participants rated each statement on a 5-point Likert scale (1~=~Strongly disagree, 5~=~Strongly agree).

\begin{itemize}
    \item The system provided choices based on my true interests.
    \item The system let me do things my own way.
    \item The system helped me take actions that I wanted to do rather than because I was told to.
    \item The system let me be in control of what I did.
    \item The system helped me make my own decisions.
\end{itemize}

\subsection{Transparency Of Robots Scale (TOROS)}
Participants rated each statement on a 7-point Likert scale (1=Strongly disagree, 7=Strongly agree).

\begin{itemize}
    \item I am confused about the robot’s general objectives.
    \item I cannot explain the robot’s behavior.
    \item It is clear to me what the robot does.
    \item I have a clear understanding of how the robot operates in general.
    \item I feel informed about the robot’s activities.
    \item The robot’s behavior is predictable.
\end{itemize}

\subsection{NASA-TLX}
Participants rated each item on a 7-point Likert scale (1=Very low, 7=Very high).

\begin{itemize}
    \item How mentally demanding was the task?
    \item How physically demanding was the task?
    \item How hurried or rushed was the pace of the task?
    \item How successful were you in accomplishing what you were asked to do?
    \item How hard did you have to work to accomplish your level of performance?
    \item How insecure, discouraged, irritated, stressed, and annoyed were you?
\end{itemize}

\subsection{System Usability Scale (SUS)}
Participants rated each statement on a 5-point Likert scale (1=Strongly disagree, 5=Strongly agree).

\begin{itemize}
    \item I think that I would like to use this system frequently.
    \item I found the system unnecessarily complex.
    \item I thought the system was easy to use.
    \item I think that I would need the support of a technical person to be able to use this system.
    \item I found the various functions in this system were well integrated.
    \item I thought there was too much inconsistency in this system.
    \item I would imagine that most people would learn to use this system very quickly.
    \item I found the system very cumbersome to use.
    \item I felt very confident using the system.
    \item I needed to learn a lot of things before I could get going with this system.
\end{itemize}

\subsection{Open-Ended Questions}

\begin{itemize}
    \item Across all three trials, how was your overall experience using the system?
    \item How would you compare your experience across the three trials?
    \item Which trial did you prefer, and what made it better than the others?
    \item Across all three trials, which interface features did you find most useful, and why?
    \item Do you have any other thoughts or feedback?
\end{itemize}

\section{Appendix: Plan Examples}
\label{sec:planExamples}

\newcommand{\appendixline}[2][0em]{\noindent\hspace*{#1}\detokenize{#2}\par}
\newcommand{\appendixhlline}[2][0em]{\noindent\hspace*{#1}\textbf{\detokenize{#2}}\par}
\newenvironment{appendixblock}{\begingroup\small\ttfamily\raggedright\sloppy\parindent0pt}{\par\endgroup}

For readability, all numeric plan parameters in this appendix are rounded to at most two decimal places. High-level primitives are numbered as X., and the low-level primitives nested under them are numbered as X.Y. All plans are selected from task history before robot execution.\par

\subsection*{Task 1: \tOne}
\textbf{P11, Trial 3}\\

\noindent\textit{User Prompt. }
\begin{appendixblock}
\appendixline{Move bowl to the center of the table. Move the cup next to the bowl in the upper left side of the bowl. Using the Right Arm of the Robot, move the spoon 5 inches from the center of the table.}
\end{appendixblock}

\subsection*{Task 2: \tTwo}
\textbf{P37, Trial 3} \\

\noindent\textit{User Prompt. }
\begin{appendixblock}
\appendixline{Pick up a cup and empty its contents into the bowl. Place the cup upright on the table at its original position. Repeat all actions with the second cup}
\end{appendixblock}

\subsection*{Task 3: \tThree}
\textbf{P13, Trial2 } \\
\noindent\textit{Prompt. }
\begin{appendixblock}
\appendixline{There are four objects on the table, 3 to be cleaned and a fourth to store the items.  The items to tidy are two cups and a bowl.  The large bin will collect the items.  The goal is to place the two cups and bowl in the large bin.}
\end{appendixblock}

\input{tables/plan_task1_example}
\input{tables/plan_task3_example}
\input{tables/plan_task2_example}
\clearpage

\section{Appendix: Additional Data Visualizations}
\label{sec:extra_figures}

\begin{figure}[H]
  \includegraphics[width=\columnwidth]{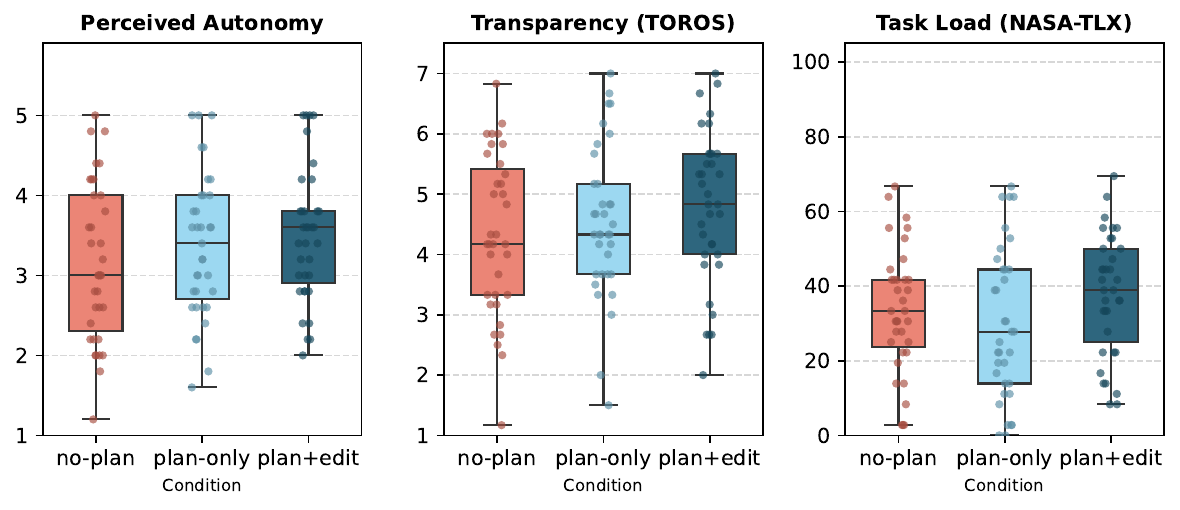}
  \caption{ \textit{Distribution of User Experience Measures} --- This shows the distribution of data for participants score for Perceived Autonomy (control), Transparency, and Task Load. The box plot shows the inter quartile range (IQR) Q1-Q3 with the middle horizontal line showing the mean. The boxplot whiskers show 1.5x IQR. The boxplot is overlaid with dots representing the individual participant scores.}
  \label{fig:box_plot}
\end{figure}

\begin{figure}[H]
  \includegraphics[width=\columnwidth]{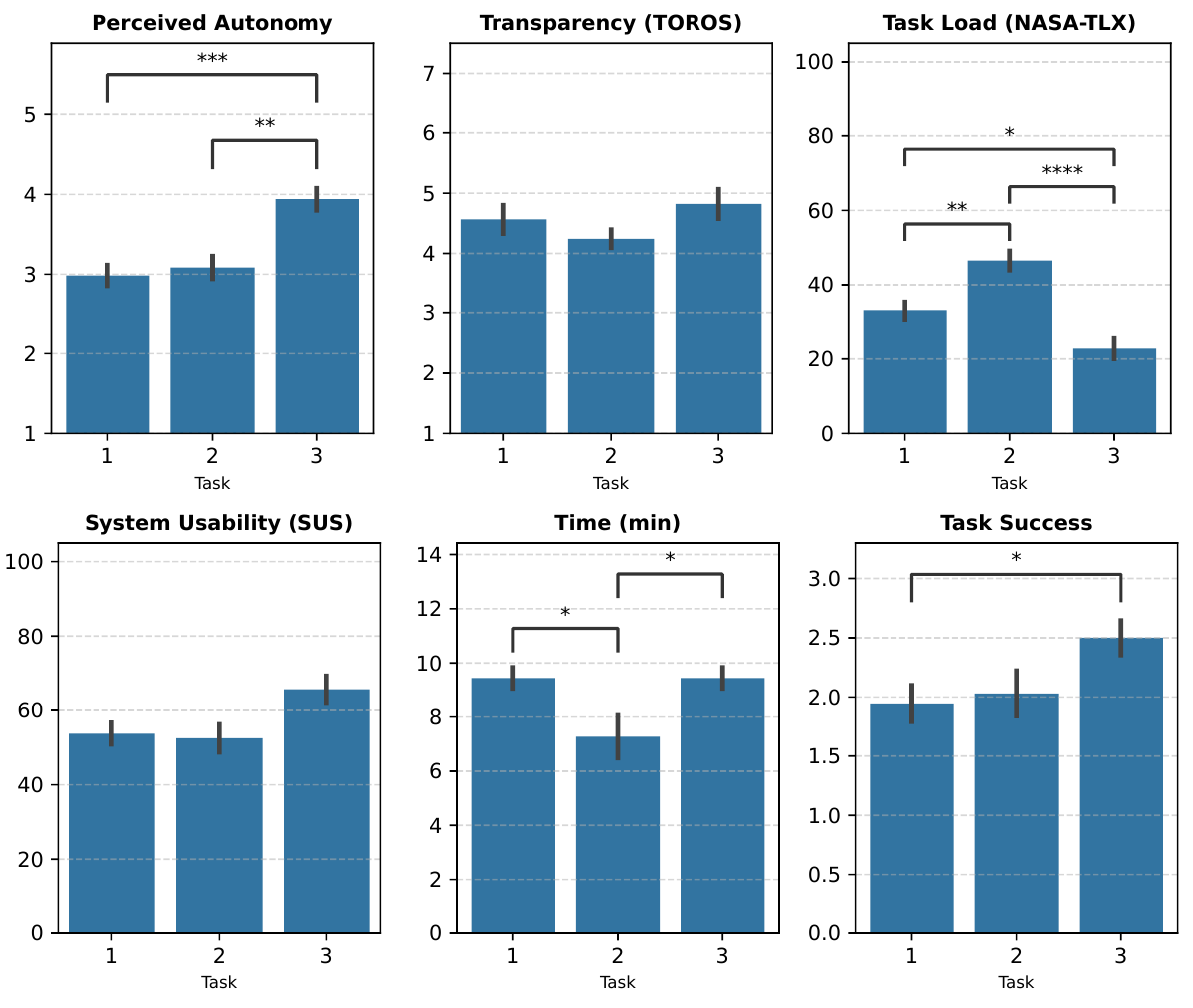}
  \caption{\textit{Quantitative Measures Across Tasks} --- The bar chart shows the mean scores across tasks for participants scores for Perceived Autonomy (control), Transparency, Task Load, System Usability, and mean measures for Time (minutes) and Task Success (out of 3). For Perceived Autonomy Transparency, System Usability, and Task Success a higher score is better, while for Task Load and Time a lower value is better. Task 1 is \tOne, 2 is \tTwo, and 3 is \tThree. Horizontal brackets indicate significant pairwise comparison, calculated with a LMM $(p < 0.05^{*},\ p < 0.01^{**},\ p < 0.001^{***})$. Vertical lines on each bar indicate standard error. }
  \label{fig:mean_by_only_task_full}
\end{figure}

\begin{figure*}[b]
  \includegraphics[width=\textwidth]{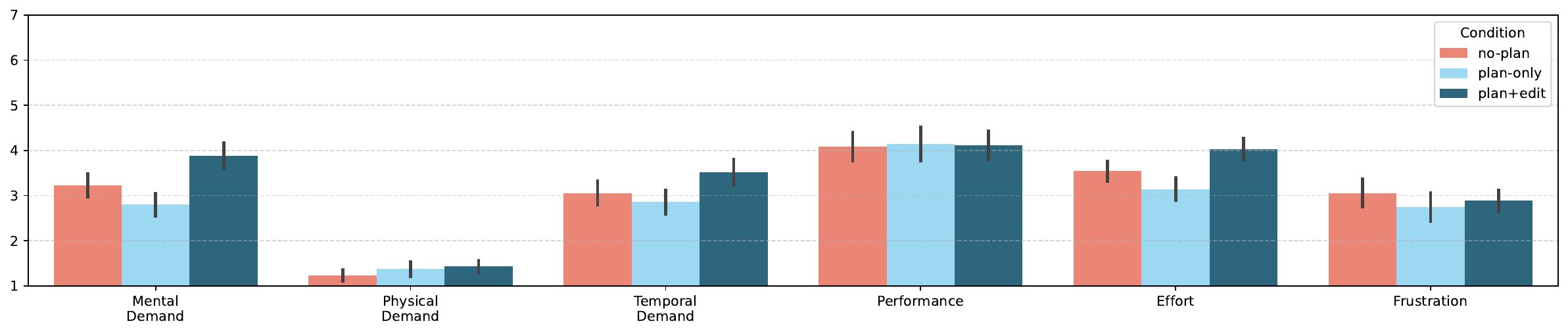}
  \caption{ \textit{Task Load Scores By Question} --- The bar chart shows the mean participant scores for Task Load (NASA-TLX) per questions. For all questions, a lower score is better.}
  \label{fig:nasa_tlx_qs}
\end{figure*}

Here we present Figures \ref{fig:box_plot}, \ref{fig:mean_by_only_task_full}, and \ref{fig:nasa_tlx_qs} as supplementary data visualizations.

\end{document}

%% file: tables/primitives.tex
\begin{table*}[b]
\centering
\small
\begin{tabular}{llp{5.5cm}p{4.8cm}p{3cm}}
\toprule
\textbf{Level} & \textbf{Name} & \textbf{Description} & \textbf{Parameters} & \textbf{Low-Level Sequence}\\
\midrule

\multirow{6}{*}{Low}
& home & Move both arms to their home position & -- & --\\
& move\_to\_pose & Move the end effector to a specified pose & arm, pose, object & --\\
& grasp & Close the gripper & arm, object & --\\
& release & Open the gripper & arm, object & --\\
& tilt\_in\_hand & Tilt object in hand & arm, object & --\\
& wait & Pause both arms for a duration & duration & --\\

\midrule

\multirow{3}{*}{High}
& pick & Pick up an object to move it to a target position & arm, grasp\_pose, end\_position, object
    & move\_to\_pose(x2), grasp, move\_to\_pose(x3) \\
& pour & Tilt to pour, hold, then return to initial pose & arm, initial\_pose, pour\_orientation, pour\_hold, pour\_object, receiving\_object
    & move\_to\_pose, tilt\_in\_hand, move\_to\_pose, wait, move\_to\_pose \\
& pick\_and\_place & Pick up an object and place it at a target position & arm, grasp\_pose, end\_position, object 
    & move\_to\_pose(x2), grasp, move\_to\_pose(x4), release, move\_to\_pose\\

\bottomrule
\end{tabular}
\caption{\textit{Primitives} --- The robot plan consists of robot primitives. Low-level primitives are building block robot actions. High-level robot primitives are made up of sequences of low-level primitives. Each primitive has a set of parameters which customize the primitive execution.}
\label{tab:primitives}
\end{table*}

%% file: tables/tasks.tex
\begin{table}[b]
\small
\centering
\begin{tabularx}{\columnwidth}{p{0.1cm} p{2.1cm} p{5.8cm}}
\toprule
\textbf{\#} & \textbf{Task} & \textbf{Subtask Success Criteria} \\
\midrule

1 & \tOne\newline (bowl, cup, spoon)
  & (1) Bowl placed in the center of the table \\
  & 
  & (2) Cup placed at upper left of bowl within 10cm \\
  & 
  & (3) Spoon placed at right of bowl within 10cm \\

\midrule

2 & \tTwo\newline (cup$\times$2, bowl)
  & (1) First cup poured into bowl \\
  & 
  & (2) First cup moved out of the way of the bowl \\
  & 
  & (3) Second cup poured into the bowl \\

\midrule

3 & \tThree\newline (cup$\times$2, bowl, box)
  & (1) First cup placed into the box \\
  & 
  & (2) Second cup placed into the box \\
  & 
  & (3) Bowl placed into the box \\

\bottomrule
\end{tabularx}
\caption{\textit{Experiment Tasks} --- Participants were assigned one of three tasks to complete during the experiment. Objects used in each task are listed in parentheses. For each task, binary subtask success is computed and summed to determine overall task success. }
\label{tab:tasks}
\end{table}

%% file: tables/iteration.tex
\begin{table*}[t]
\centering
\begin{tabular}{p{1.4cm} p{3.3cm} p{3.3cm} p{3.5cm} p{3.8cm}}
\toprule
\textbf{Iteration Strategies}  & \textbf{Iteration 1} & \textbf{Iteration 2} & \textbf{Iteration 3} & \textbf{Iteration 4} \\
\midrule

\sOne 
& Move the cup into the box 
& Move the other cup into the box 
& Move the red bowl into the box 
& Move two cups and the red bowl into the box \\
\midrule

\sTwo
& move cup right, move cup forward and pour into bowl. 
& move cup right, move cup forward and pour into bowl. \textbf{move cup left and place on table }
& move cup right, move cup forward and pour into bowl. move cup left and place on table. \textbf{grab second cup, move right, move forward and pour into bowl. move cup left and place on table.} 
&  \\
\midrule

\sThree
& Move bowl to the center of the table. Move the cup to the upper left from the bowl. Using the Right Arm of the Robot, move the spoon 5 inches from the bowl. 
& Move bowl to the center of the table. Move the cup \textbf{next to the bowl in the upper left side of the bowl}. Using the Right Arm of the Robot, move the spoon 5 inches from the bowl. 
& Move bowl to the center of the table. Move the cup next to the bowl in the upper left side of the bowl. Using the Right Arm of the Robot,\textbf{ move the spoon 5 inches from the center of the table}. 
& Move bowl to the center of the table. Using the Right Arm of the Robot, move the spoon 5 inches from the center of the table. \textbf{Using the Left Arm of the Robot} move the cup to the upper left \textbf{corner} of the bowl. \\

\bottomrule

\end{tabular}
\caption{\textit{Iteration Strategies} --- During the user study, participants used three types of prompt iteration strategies. Here we show examples of these strategies from P10, P12, and P11 respectively. Textual changes are shown in bold.}
\label{tab:iteration}
\end{table*}

%% file: tables/plan_task1_example.tex
\begin{table*}[t]
\centering
\scriptsize
\begin{tabular*}{\textwidth}{@{\extracolsep{\fill}} c l c c l}
\hline
\multirow{2}{*}{Sequence} & \multirow{2}{*}{Primitive} & \multicolumn{3}{c}{Parameters} \\
\cline{3-5}
 & & Arm & Object & Pose \\
\hline

\multicolumn{5}{l}{\textbf{1. home}} \\

\hline
\multicolumn{5}{l}{\textbf{2. pick\_and\_place (bowl\_1)}} \\
\multicolumn{5}{l}{arm=left,\ grasp\_pose=[-0.08,0.12,1.01,1,0,0,0],\ end\_position=[0,0,1.03]} \\

2.1 & move\_to\_pose & left  & --      & [-0.08, 0.12, 1.13, 1, 0, 0, 0] \\
2.2 & move\_to\_pose & left  & --      & [-0.08, 0.12, 1.01, 1, 0, 0, 0] \\
2.3 & pincer\_grasp & left  & bowl\_1 & -- \\
2.4 & move\_to\_pose & left  & bowl\_1 & [-0.08, 0.12, 1.13, 1, 0, 0, 0] \\
2.5 & move\_to\_pose & right & --      & [0.4, 0, 1.4, 1, 0, 0, 0] \\
2.6 & move\_to\_pose & left  & bowl\_1 & [0.0, 0.0, 1.13, 1, 0, 0, 0] \\
2.7 & move\_to\_pose & left  & bowl\_1 & [0.0, 0.0, 1.03, 1, 0, 0, 0] \\
2.8 & release        & left  & bowl\_1 & -- \\
2.9 & move\_to\_pose & left  & --      & [0.0, 0.0, 1.13, 1, 0, 0, 0] \\

\hline
\multicolumn{5}{l}{\textbf{3. pick\_and\_place (cup\_1)}} \\
\multicolumn{5}{l}{arm=left,\ grasp\_pose=[-0.22,0.2,1.02,0.71,0.71,0,0],\ end\_position=[-0.04,0.18,1.04]} \\

3.1 & move\_to\_pose & left & --     & [-0.22, 0.2, 1.14, 0.71, 0.71, 0, 0] \\
3.2 & move\_to\_pose & left & --     & [-0.22, 0.2, 1.02, 0.71, 0.71, 0, 0] \\
3.3 & pincer\_grasp & left & cup\_1 & -- \\
3.4 & move\_to\_pose & left & cup\_1 & [-0.22, 0.2, 1.14, 0.71, 0.71, 0, 0] \\
3.5 & move\_to\_pose & left & cup\_1 & [-0.04, 0.18, 1.14, 0.71, 0.71, 0, 0] \\
3.6 & move\_to\_pose & left & cup\_1 & [-0.04, 0.18, 1.04, 0.71, 0.71, 0, 0] \\
3.7 & release        & left & cup\_1 & -- \\
3.8 & move\_to\_pose & left & --     & [-0.04, 0.18, 1.14, 0.71, 0.71, 0, 0] \\

\hline
\multicolumn{5}{l}{\textbf{4. pick\_and\_place (spoon\_1)}} \\
\multicolumn{5}{l}{arm=right,\ grasp\_pose=[0.17,0.19,1.01,1,0,0,0],\ end\_position=[0.13,0.0,1.11]} \\

4.1 & move\_to\_pose & left  & --       & [-0.4, 0, 1.4, 1, 0, 0, 0] \\
4.2 & move\_to\_pose & right & --       & [0.17, 0.19, 1.21, 1, 0, 0, 0] \\
4.3 & move\_to\_pose & right & --       & [0.17, 0.19, 1.01, 1, 0, 0, 0] \\
4.4 & pincer\_grasp & right & spoon\_1 & -- \\
4.5 & move\_to\_pose & right & spoon\_1 & [0.17, 0.19, 1.21, 1, 0, 0, 0] \\
4.6 & move\_to\_pose & right & spoon\_1 & [0.13, 0.0, 1.21, 1, 0, 0, 0] \\
4.7 & move\_to\_pose & right & spoon\_1 & [0.13, 0.0, 1.11, 1, 0, 0, 0] \\
4.8 & release        & right & spoon\_1 & -- \\
4.9 & move\_to\_pose & right & --       & [0.13, 0.0, 1.21, 1, 0, 0, 0] \\

\hline
\end{tabular*}
\label{tab:task1_plan}
\caption{Plan for \tOne\ from P11, Trial 3. Includes high-level and low-level primitives.}
\end{table*}

%% file: tables/plan_task3_example.tex
\begin{table*}[t]
\centering
\scriptsize
\begin{tabular*}{\textwidth}{@{\extracolsep{\fill}} c l c c l}
\hline
\multirow{2}{*}{Sequence} & \multirow{2}{*}{Primitive} & \multicolumn{3}{c}{Parameters} \\
\cline{3-5}
 & & Arm & Object & Pose \\
\hline

\multicolumn{5}{l}{\textbf{1. home}} \\

\hline
\multicolumn{5}{l}{\textbf{2. pick\_and\_place (cup\_1)}} \\
\multicolumn{5}{l}{arm=right,\ grasp\_pose=[0.16,0.24,1.02,0.71,0.71,0,0],\ end\_position=[0.0,-0.15,1.21]} \\

2.1 & move\_to\_pose & right & --     & [0.16, 0.24, 1.32, 0.71, 0.71, 0, 0] \\
2.2 & move\_to\_pose & right & --     & [0.16, 0.24, 1.02, 0.71, 0.71, 0, 0] \\
2.3 & pincer\_grasp & right & cup\_1 & -- \\
2.4 & move\_to\_pose & right & cup\_1 & [0.16, 0.24, 1.32, 0.71, 0.71, 0, 0] \\
2.5 & move\_to\_pose & left  & --     & [-0.4, 0, 1.4, 1, 0, 0, 0] \\
2.6 & move\_to\_pose & right & cup\_1 & [0.0, -0.15, 1.32, 0.71, 0.71, 0, 0] \\
2.7 & move\_to\_pose & right & cup\_1 & [0.0, -0.15, 1.21, 0.71, 0.71, 0, 0] \\
2.8 & release        & right & cup\_1 & -- \\
2.9 & move\_to\_pose & right & --     & [0.0, -0.15, 1.32, 0.71, 0.71, 0, 0] \\

\hline
\multicolumn{5}{l}{\textbf{3. pick\_and\_place (cup\_2)}} \\
\multicolumn{5}{l}{arm=right,\ grasp\_pose=[0.03,0.18,1.02,0.71,0.71,0,0],\ end\_position=[0.0,-0.15,1.21]} \\

3.1 & move\_to\_pose & right & --     & [0.03, 0.18, 1.32, 0.71, 0.71, 0, 0] \\
3.2 & move\_to\_pose & right & --     & [0.03, 0.18, 1.02, 0.71, 0.71, 0, 0] \\
3.3 & pincer\_grasp & right & cup\_2 & -- \\
3.4 & move\_to\_pose & right & cup\_2 & [0.03, 0.18, 1.32, 0.71, 0.71, 0, 0] \\
3.5 & move\_to\_pose & right & cup\_2 & [0.0, -0.15, 1.32, 0.71, 0.71, 0, 0] \\
3.6 & move\_to\_pose & right & cup\_2 & [0.0, -0.15, 1.21, 0.71, 0.71, 0, 0] \\
3.7 & release        & right & cup\_2 & -- \\
3.8 & move\_to\_pose & right & --     & [0.0, -0.15, 1.32, 0.71, 0.71, 0, 0] \\

\hline
\multicolumn{5}{l}{\textbf{4. pick\_and\_place (bowl\_1)}} \\
\multicolumn{5}{l}{arm=left,\ grasp\_pose=[-0.19,0.1,1.01,1,0,0,0],\ end\_position=[0.0,-0.15,1.2]} \\

4.1 & move\_to\_pose & right & --       & [0.4, 0, 1.4, 1, 0, 0, 0] \\
4.2 & move\_to\_pose & left  & --       & [-0.19, 0.1, 1.3, 1, 0, 0, 0] \\
4.3 & move\_to\_pose & left  & --       & [-0.19, 0.1, 1.01, 1, 0, 0, 0] \\
4.4 & pincer\_grasp & left  & bowl\_1 & -- \\
4.5 & move\_to\_pose & left  & bowl\_1 & [-0.19, 0.1, 1.3, 1, 0, 0, 0] \\
4.6 & move\_to\_pose & left  & bowl\_1 & [0.0, -0.15, 1.3, 1, 0, 0, 0] \\
4.7 & move\_to\_pose & left  & bowl\_1 & [0.0, -0.15, 1.2, 1, 0, 0, 0] \\
4.8 & release        & left  & bowl\_1 & -- \\
4.9 & move\_to\_pose & left  & --       & [0.0, -0.15, 1.3, 1, 0, 0, 0] \\

\hline
\end{tabular*}
\label{tab:task3_plan}
\caption{Plan for \tThree\ from P13, Trial 2. Includes high-level and low-level primitives.}
\end{table*}

%% file: tables/plan_task2_example.tex
\begin{table*}[t]
\centering
\scriptsize
\begin{tabular*}{\textwidth}{@{\extracolsep{\fill}} c l c c l}
\hline
\multirow{2}{*}{Sequence} & \multirow{2}{*}{Primitive} & \multicolumn{3}{c}{Parameters} \\
\cline{3-5}
 & & Arm & Object & Pose \\
\hline

\multicolumn{5}{l}{\textbf{1. home}} \\
1 & home & -- & -- & -- \\

\hline
\multicolumn{5}{l}{\textbf{2. pick (cup\_1)}} \\
\multicolumn{5}{l}{arm=left,\ grasp\_pose=[-0.1,0.19,1.02,0.71,0.71,0,0],\ end\_position=[-0.1,0.19,1.04]} \\

2.1 & move\_to\_pose & left & --     & [-0.1, 0.19, 1.14, 0.71, 0.71, 0, 0] \\
2.2 & move\_to\_pose & left & --     & [-0.1, 0.19, 1.02, 0.71, 0.71, 0, 0] \\
2.3 & pincer\_grasp & left & cup\_1 & -- \\
2.4 & move\_to\_pose & left & cup\_1 & [-0.1, 0.19, 1.14, 0.71, 0.71, 0, 0] \\
2.5 & move\_to\_pose & left & cup\_1 & [-0.1, 0.19, 1.14, 0.71, 0.71, 0, 0] \\
2.6 & move\_to\_pose & left & cup\_1 & [-0.1, 0.19, 1.04, 0.71, 0.71, 0, 0] \\

\hline
\multicolumn{5}{l}{\textbf{3. pour (cup\_1 $\rightarrow$ bowl\_1)}} \\
\multicolumn{5}{l}{arm=left,\ initial\_pose=[0.01,-0.09,1.1,1,0,0,0],\ pour\_orientation=[-0.57,0,0,0.82],\ pour\_hold=1.5} \\

3.1 & move\_to\_pose & left & cup\_1 & [0.01, -0.09, 1.1, 1, 0, 0, 0] \\
3.2 & tilt\_in\_hand & left & cup\_1 & -- \\
3.3 & move\_to\_pose & left & cup\_1 & [0.01, -0.09, 1.1, -0.57, 0, 0, 0.82] \\
3.4 & wait           & --   & --     & duration=1.5 \\
3.5 & move\_to\_pose & left & cup\_1 & [0.01, -0.09, 1.1, 1, 0, 0, 0] \\

\hline
\multicolumn{5}{l}{\textbf{4. move\_to\_pose (cup\_1)}} \\
 & move\_to\_pose & left & cup\_1 & [-0.1, 0.19, 1.1, 1, 0, 0, 0] \\

\hline
\multicolumn{5}{l}{\textbf{5. release (cup\_1)}} \\
& release & left & cup\_1 & -- \\

\hline
\multicolumn{5}{l}{\textbf{6. pick (cup\_2)}} \\
\multicolumn{5}{l}{arm=left,\ grasp\_pose=[-0.22,0.2,1.02,0.71,0.71,0,0],\ end\_position=[-0.22,0.2,1.04]} \\

6.1 & move\_to\_pose & left & --     & [-0.22, 0.2, 1.14, 0.71, 0.71, 0, 0] \\
6.2 & move\_to\_pose & left & --     & [-0.22, 0.2, 1.02, 0.71, 0.71, 0, 0] \\
6.3 & pincer\_grasp & left & cup\_2 & -- \\
6.4 & move\_to\_pose & left & cup\_2 & [-0.22, 0.2, 1.14, 0.71, 0.71, 0, 0] \\
6.5 & move\_to\_pose & left & cup\_2 & [-0.22, 0.2, 1.14, 0.71, 0.71, 0, 0] \\
6.6 & move\_to\_pose & left & cup\_2 & [-0.22, 0.2, 1.04, 0.71, 0.71, 0, 0] \\

\hline
\multicolumn{5}{l}{\textbf{7. pour (cup\_2 $\rightarrow$ bowl\_1)}} \\
\multicolumn{5}{l}{arm=left,\ initial\_pose=[0.01,-0.09,1.21,1,0,0,0],\ pour\_orientation=[-0.57,0,0,0.82],\ pour\_hold=1.5} \\

7.1 & move\_to\_pose & right & --     & [0.4, 0, 1.4, 1, 0, 0, 0] \\
7.2 & move\_to\_pose & left  & cup\_2 & [0.01, -0.09, 1.21, 1, 0, 0, 0] \\
7.3 & tilt\_in\_hand & left  & cup\_2 & -- \\
7.4 & move\_to\_pose & left  & cup\_2 & [0.01, -0.09, 1.21, -0.57, 0, 0, 0.82] \\
7.5 & wait           & --    & --     & duration=1.5 \\
7.6 & move\_to\_pose & left  & cup\_2 & [0.01, -0.09, 1.21, 1, 0, 0, 0] \\

\hline
\multicolumn{5}{l}{\textbf{8. move\_to\_pose (cup\_2)}} \\
& move\_to\_pose & left & cup\_2 & [-0.22, 0.2, 1.04, 1, 0, 0, 0] \\

\hline
\multicolumn{5}{l}{\textbf{9. release (cup\_2)}} \\
& release & left & cup\_2 & -- \\

\hline
\end{tabular*}
\label{tab:task2_plan}
\caption{Plan for \tTwo\ from P37, Trial 3. Includes high-level and low-level primitives.}\end{table*}